\documentclass[runningheads]{llncs}

\usepackage[utf8]{inputenc}
\usepackage{amsfonts}
\usepackage{amsmath}
\usepackage{subcaption}
\usepackage{comment}
\usepackage{graphicx}
\graphicspath{ {./images/} }
\usepackage[ruled,vlined,linesnumbered]{algorithm2e}
\usepackage{cite}

\newcommand{\netpolicy}{\ensuremath{\pi(a|s)}}  
\newcommand{\mbf}[1]{\mathbf{#1}}  
\newcommand{\mc}[1]{\mathcal{#1}}
\newcommand{\R}{\mathbb{R}}
\newcommand{\ra}{\rightarrow}

\newcommand\blfootnote[1]{%
  \begingroup
  \renewcommand\thefootnote{}\footnote{#1}%
  \addtocounter{footnote}{-1}%
  \endgroup
}

\begin{document}
\title{Controlling Level of Unconsciousness by Titrating Propofol with Deep Reinforcement Learning}
\titlerunning{Controlling Level of Unconsciousness with Deep Reinforcement Learning}

\author{Gabriel Schamberg$^\star$\inst{1,2} 
\and
Marcus Badgeley$^\star$\inst{2,3}
\and
Emery N. Brown\inst{1,2,3}}
\authorrunning{G. Schamberg et al.}

\institute{
Picower Institute for Learning and Memory, Massachusetts Institute of Technology, Cambridge, MA 02139 
\and
Department of Brain and Cognitive Sciences, Massachusetts Institute of Technology, Cambridge, MA 02139
\and
Department of Anesthesiology, Critical Care and Pain Medicine, Massachusetts General Hospital, Boston, MA 02114
}

\maketitle              

\blfootnote{This work was supported by Picower Postdoctoral Fellowship (to GS) and the National Institutes of Health P01 GM118629 (to ENB).}

\blfootnote{$^\star$These authors contributed equally to this work.}

\begin{abstract}
Reinforcement Learning (RL) can be used to fit a mapping from patient state to a medication regimen.  Prior studies have used deterministic and value-based tabular learning to learn a propofol dose from an observed anesthetic state. Deep RL replaces the table with a deep neural network and has been used to learn medication regimens from registry databases. Here we perform the first application of deep RL to closed-loop control of anesthetic dosing in a simulated environment.  We use the cross-entropy method to train a deep neural network to map an observed anesthetic state to a probability of infusing a fixed propofol dosage. During testing, we implement a deterministic policy that transforms the probability of infusion to a continuous infusion rate. The model is trained and tested on simulated pharmacokinetic/pharmacodynamic models with randomized parameters to ensure robustness to patient variability. The deep RL agent significantly outperformed a proportional-integral-derivative controller (median absolute performance error 1.7\% $\pm$ {0.6} and 3.4\% $\pm$ {1.2}). Modeling continuous input variables instead of a table affords more robust pattern recognition and utilizes our prior domain knowledge. Deep RL learned a smooth policy with a natural interpretation to data scientists and anesthesia care providers alike.
\keywords{anesthesia  \and reinforcement learning \and deep learning}
\end{abstract}

\section{Introduction}

The proliferation of anesthesia in the 1800s is America's greatest contribution to modern medicine and enabled far more complex, invasive, and humane surgical procedures. Now nearly 60,000 patients receive general anesthesia for surgery daily in the United States \cite{brown2010general}. Anesthesia is a reversible drug-induced state characterized by a combination of amnesia, immobility, antinociception, and loss of consciousness \cite{brown2010general}. Anesthesia providers are not only responsible for a patient's depth of anesthesia, but also their physiologic stability and oxygen delivery.

Anesthesiologists need to determine the medication regimen a patient receives throughout a surgical procedure. The anesthetic state is managed by providing inhaled vapors or infusing intravenous medication. The medication most studied for controlling the patient's level of unconsciousness is propofol. Propofol affects the brain's cortex and arousal centers to induce loss of consciousness in a dose-dependent manner.  Propofol dosage needs to be balanced: patients should be deep enough to avoid intraoperative awareness, but too much anesthesia can cause physiologic instability or cognitive deficits.  Currently anesthesiologists can manually calculate and inject each dose, or select the desired concentration of propofol in the brain, and an infusion pump will adjust infusion rates based on how an average patient processes the medication.

Investigational devices and studies have shown that measuring brain activity can provide personalized computer-calculated dosing regimens. Studies of automatic anesthetic administration have three primary components. First, \emph{sensing} involves automatically obtaining a numerical representation of the patient's anesthetic state. Prior studies have primarily focused on controlling the level of unconsciousness (LoU) using a variety of indices, including the bispectral index (BIS) \cite{gentilini2001modeling,absalom2002closed}, $WAV_{CNS}$ \cite{dumont2009robust}, and burst suppression probability \cite{westover2015robust,shanechi2013brain}. Second, \emph{modeling} involves the development of pharmacokinetic (PK) and pharmacodynamic (PD) models of how a patient's LoU responds to specified drug dosages. These models are used to derive optimal control laws \cite{shanechi2013brain}, tune controller parameters \cite{westover2015robust}, and/or develop robust controller parameterizations \cite{dumont2009robust}. Finally, the \emph{controller} determines the mapping from sensed variables to drug infusion rates. Numerous control algorithms have been studied for LoU regulation, including (but not limited to) proportional integral derivative (PID) controllers, model predictive (MP) controllers, and linear quadratic regulator (LQR) controllers. The performance of these algorithms is restricted by linearity assumptions and/or reliance on a nominal patient model for obtaining the control action.

Reinforcement learning (RL) is a form of optimal control which learns by optimizing a flexible reward system. RL can be used to fit a mapping from anesthetic state to a propofol dose. Contrary to MP and LQR controllers, the RL-based controllers can be \emph{model-free} in that the control law is established without any knowledge of the underlying model. Prior studies using tabular RL created a table where each entry represents a discrete propofol dosage that corresponds to a discrete observation \cite{moore2011reinforcement, moore2014reinforcement,padmanabhan2015closed}. Tabular mappings are flexible but do not scale well with larger state spaces and can have non-smooth policies that result from independently determining actions for each of the discretized states. Continuous states have been used by actor-critic methods that use linear function maps \cite{lowery2013towards} and adaptive linear control  \cite{padmanabhan2019optimal}.  Existing studies that train RL agents to administer anesthesia tend to either underconstrain by disregarding continuity or overconstrain by imposing strict linearity assumptions.

The use of deep neural networks as functional maps in RL (called ``deep RL'') has been used to learn medication regimens from registry databases outside of the anesthesia context \cite{prasad2017reinforcement, lopezmartinez2019deeprlanalgesia}. Existing deep RL studies use large observation spaces and disregard known PK/PD properties. By using retrospectively collected data, these studies do not permit the RL agent to learn from its own actions and restrict the agent's ``teachable moments'' to those that are observed in the data.  

We perform the first application of deep RL to anesthetic dosing supported with fundamental models from pharmacology. Using data from a simulated PK/PD model, we implement an RL framework for training a neural network to provide a mapping from a continuous valued observation vector to a distribution over actions. The use of a deep neural network allows the number of parameters of the model to scale linearly with the number of inputs to the policy map, avoiding the exponential growth that occurs when expanding the input dimension of tabular policies. The resultant policy can represent nonlinear functions while still yielding a smooth function of the input variables. By training the RL agent on simulated data, we can control the range of patient models that are included in the learning process. As such, the proposed framework allows us to experiment with a variety of policy inputs in order to directly incorporate robustness to patient variability into the training procedure.

\section{Methods}

\begin{figure}[t]
    \centering
    \includegraphics[width=0.8\textwidth]{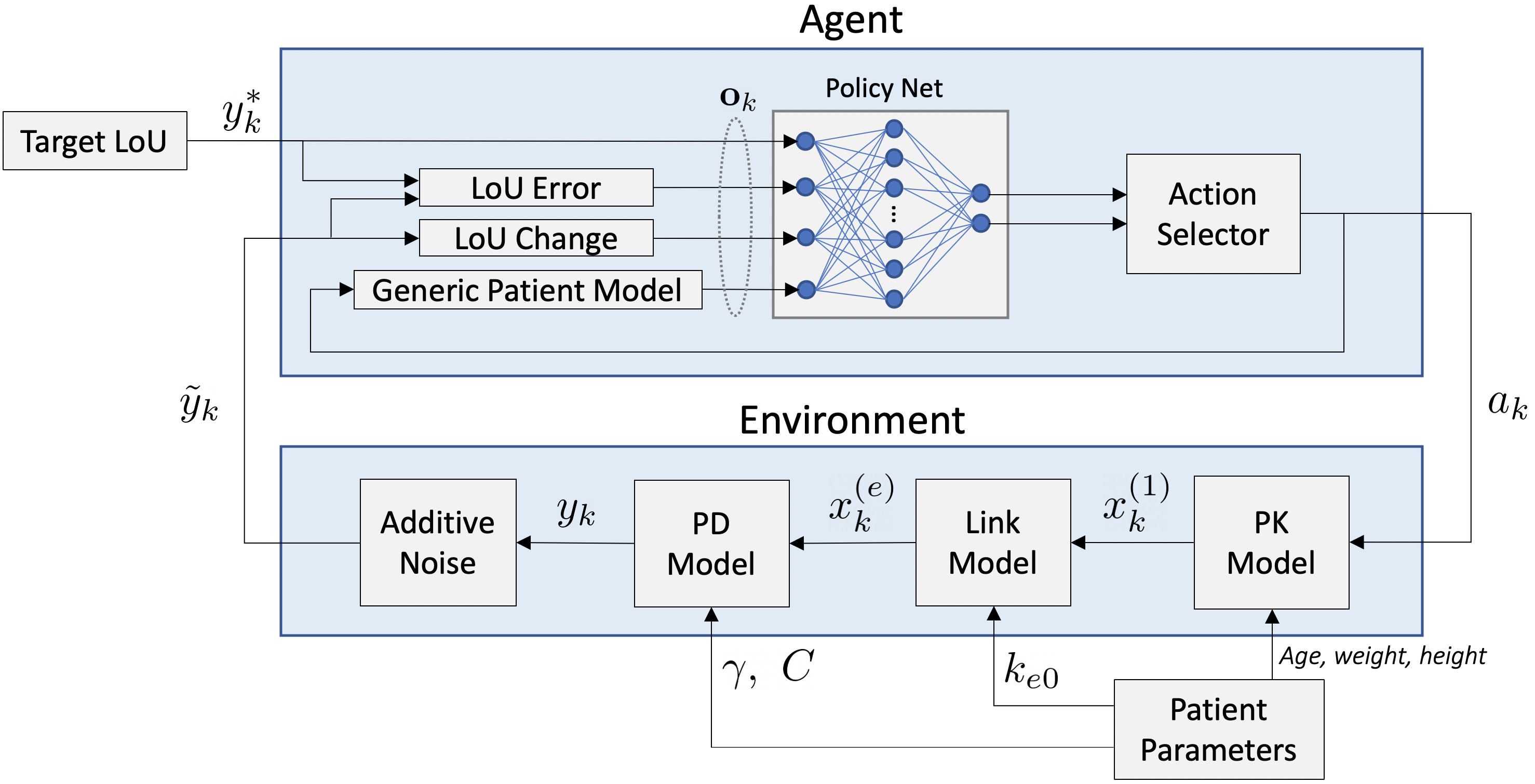}
    \caption{Block diagram of the proposed RL framework.}        \label{fig:diagram}
\end{figure}

In this section we develop a mathematical formalization for learning how to administer propofol to control LoU. We formalize this propofol dosing task as a partially observable Markov decision process (POMDP) and solve it using the RL method cross-entropy. This RL ``agent'' learns from data generated by simulated interactions with the environment PK/PD state-space model (see Figure \ref{fig:diagram}).

\subsection{Environment Model}
The primary component of the environment is the patient model, which dictates the observed LoU given a drug infusion profile and is composed of three sub-models. First, a discrete time 3-compartment pharmacokinetic (PK) model is used to model the mass transfer of infused drug between the central, slow peripheral, and rapid peripheral compartments:
\begin{equation}\label{eq:pk}
    \mbf{x}_{k+1} = \mbf{A}\mbf{x}_k + \mbf{B}a_k
\end{equation}
where $\mbf{x}_k=[x^{(1)}_k,x^{(2)}_k,x^{(3)}_k] \in \R^3_+$ represents the 3-compartment model concentrations (where $\R_+$ represents the set of non-negative reals), $a_k \in\mc{A}=\{0,1\}$ represents whether or not drug is infused at time $k$, $\mbf{A}\in \R^{3\times 3}$ gives the mass transfer rates between compartments, and $\mbf{B}=[\Delta u,0,0]$ represents the mass transfer rate resulting from drug infusion. The parameters of $\mbf{A}$ are determined by the patient age, height, and weight according to the Schnider model \cite{schnider1998influence}. In the current study we have $\Delta=5$ sec (as in \cite{moore2011reinforcement}) and $u=1.67$ mg/s, such that over each five second window is either 0 mg or 8.35 mg of propofol is delivered.

The link function determines the effect site (i.e. brain) concentration from the central compartment concentration:
\begin{equation}
    x^{(e)}_{k+1} = \alpha x^{(e)}_k + \beta x^{(1)}_k
\end{equation}
where $\alpha=\exp(-k_{e0}\Delta/60)$ gives the persistence of drug in the effect site and $\beta=(k_{e0}/60)\exp(-k_{e0}\Delta/60)$ gives the mass transfer rate from the plasma compartment to the effect site for a given plasma-brain equilibration constant $k_{e0}$. 

Finally, we compute how a given effect site concentration of propofol affects LoU using a hill function:

\begin{equation}
    y_k = h(x^{(e)}_k)=\frac{{x^{(e)}_k}^\gamma}{C^\gamma + {x^{(e)}_k}^\gamma}
\end{equation}
where $y_k\in[0,1]$ gives the true LoU at time $k$, $C\in\R_+$ give the effect site concentration corresponding to a LoU of 0.5, and $\gamma\in\R_+$ determines the shape of the non-linear PD response (higher values give rise to more rapid transitions in LoU). Given that this PD model has been used in studies targeting control of BIS \cite{gentilini2001modeling}, $WAV_{CNS}$ \cite{dumont2009robust}, and BSP \cite{westover2015robust}, we choose to treat $y_k$ as a general index of LoU in the present simulation study. As such, $y_k$ can be viewed as representing a non-linear continuum from consciousness (0) to brain death (1).  The clinical interpretations of intermediate LoUs depend on the specific choice of index.

The observed LoU $\tilde{y}_k$ is obtained by adding Gaussian measurement noise $v_k\sim \mc{N}(0,\sigma_v^2)$ to the true LoU $y_k$. The resulting observed LoU is clipped to be between zero and one. We set measurement noise at $\sigma_v^2=0.0003$ \cite{moore2011reinforcement}.

\subsection{Agent Model}
At each timestep, the agent receives a measured LoU $\tilde{y}_k$ from the environment along with a target LoU $y^*_k$ and decides how much propofol to infuse. The agent uses these inputs and the infusion history to derive an observation vector: $\mathbf{o}_k=[o^{(1)}_k,\dots,o^{(d)}_k]\in\mc{O}= \R^d$. We use $d=4$ with the following observed variables: measured LoU error $o^{(1)}_k=\tilde{y}_k-y_k^*$, 30-seconds ahead predicted change in effect site concentration $o^{(2)}_k=\hat{x}^{(e)}_{k+6}-\hat{x}^{(e)}_{k}$, 30-second historical change in measured LoU $o^{(3)}_k=\tilde{y}_k-\tilde{y}_{k-6}$, and the target LoU $o^{(4)}_k=y^*_k$. The estimated effect site concentration is computed by maintaining a PK model with generic parameterization (according to Table \ref{tab:params}) throughout a trial to estimate concentration levels $\hat{\mbf{x}}_k$ and  $\hat{x}^{(e)}_{k}$. At each time, this model is propagated 30 seconds forward under the assumption that there will be no further infusion to yield $\hat{x}^{(e)}_{k+6}$, which is used to compute the predicted change. All elements of the observation vector can be computed solely from previous actions and measured LoUs, and they do not assume any knowledge of the specific patient variables in the environment. Each of the observation variables is presumed to provide the agent with a unique advantage in selecting an action. For example, the predicted change in effect site concentration is included to account for the lag between drug administration and arrival at the effect site and including the target LoU enables the agent to have different steady-state infusion rates for different target LoUs.

\begin{table}
\begin{center}
\begin{tabular}{|c|c|c|c|c|c|}
     \hline
     Sub-model & Parameter & Units & Generic & Minimum & Maximum  \\
     \hline
     PK & Height & cm& 170 & 160 & 190 \\
     PK & Weight & kg & 70 & 50 & 100 \\
     PK & Age & yr & 30 & 18 & 90 \\
     Link & $k_{e0}$ & min$^{-1}$ & 0.17 & 0.128 & 0.213 \\
     PD & $\gamma$ & - & 5 & 5 & 9 \\
     PD & $C$ & - & 2.5 & 2 & 6 \\
     \hline
\end{tabular}
\end{center}
\caption{Model parameters for the generic patient and range of parameters selected randomly for training and testing.}
\label{tab:params}
\end{table}

The agent uses a neural network to map observations to distributions over the action space. This mapping, known as the policy, is represented by the function $\pi(a_k\mid \mbf{o}_k)$, which assigns a probability to an action $a_k$ given an observation $\mbf{o}_k$. The network contains a single hidden layer with 128 nodes and ReLU activation functions, with two output nodes passed through a softmax to obtain action probabilities. The network is fully parameterized by the weights $w_\pi \in\R^{898}$ ($(4+1)\times 128 + (128+1)\times 2=898$, where the $+1$ accounts for a constant offset).

To promote exploration during training, the agent randomly selects an action according to its policy. During testing, we employ three action selection modes. In \emph{stochastic} mode, the agent randomly selects an action according to its policy as it does in training: $a^{(s)}_k \sim \pi(\cdot\mid\mbf{o}_k)$. In \emph{deterministic} mode, the agent selects the action with the highest probability: $a^{(d)}_k =\underset{a}{\operatorname{argmax}}\ \pi(a\mid\mbf{o}_k)$. In \emph{continuous} mode, the agent selects a continuous action corresponding to the policy's probability of infusing: $a^{(c)}_k =\pi(1\mid\mbf{o}_k)$. In all cases, the action yields a normalized infusion rate that is multiplied by the maximum dose $u$ in \eqref{eq:pk}.

\subsection{Cross-Entropy Training Algorithm}\label{sec:training}

The agent's policy network is trained using the cross-entropy method \cite{de2005tutorial}, an importance sampling based algorithm that is popular in training deep reinforcement learning agents \cite{szita2006learning}. The algorithm performs batches of simulations (referred to as episodes) and at the end of each batch updates the policy based on the episodes where the agent performed best. The agent's performance in a given episode is assessed using a reward function.

For a given episode $n$ in a collection of $N$ episodes, we define the episode reward to be the cumulative negative absolute error: $r_n=\sum_{k=1}^K -|y^*_{n,k}-y_{n,k}|\in(-\infty,0]$, where $K$ gives the fixed duration of an episode and $y^*_{n,k}$ and $y_{n,k}$ give the target and true LoU at time $k$ for episode $n$, respectively. After simulating $N$ episodes, the set $\mc{N}_p$ of episodes in the $p^{th}$ percentile of rewards is identified, and these are used to update the policy net parameters $w_\pi$. Letting $a_{n,k}$ be the action taken at time $k$ in episode $n$ and $\mbf{o}_{n,k}$ be the corresponding observation vector, define the cross-entropy loss for episode $n$ as:
\begin{equation}
    \mathcal{L}_n = -\sum_{k=1}^K a_{n,k} \log \pi(a_{n,k}\mid \mbf{o}_{n,k}) + (1 - a_{n,k}) \log (1 - \pi(a_{n,k}\mid \mbf{o}_{n,k}) )
\end{equation}
Given that $a_{n,k}$ is either zero or one, reducing the cross-entropy loss results in nudging the policy $\pi$ to assign a higher probability to the action $a_{n,k}$ when $\mbf{o}_{n,k}$ is observed. This nudge is accomplished by performing stochastic gradient descent (SGD) on the policy net weights with the computed losses.

In our training, we used a cutoff of $p=70\%$ (i.e. $\mc{N}_p$ includes the best 30\%) and a batch size of $N=16$. Between batches, we selected the patient parameters randomly from a uniform distribution over the parameters specified in Table \ref{tab:params}. Four LoU targets $y^*$ were sampled uniformly from $[0.25,0.75]$, with each target being used for 2,500 seconds before switching to the next, resulting in a total episode duration of 10,000 seconds, or $K=2,000$. The patient parameters and targets were kept fixed within a batch to avoid updating the policy based on the ``easiest'' environments rather than on the best performance of the agent. Further details on the training and reward system are provided in the Appendix.

\subsection{PID Controller}
Previous studies on RL-based control of propofol infusion tested control performance against a PID controller \cite{moore2011reinforcement}. The PID control action is determined by a linear combination of the instantaneous error, the error integral, and the error derivative. We use a discrete implementation of the PID controller:
\begin{equation}
    a^{(pid)}_k = K_P e_k + K_I \sum_{i=0}^k e_k + K_D \frac{e_k-e_{k-6}}{6}
\end{equation}
where $K_P$, $K_I$, and $K_D$ are the controller parameters and $e_k=y^*_k-\tilde{y}_k$ is the error. The error derivative is approximated using a 30-second lag to avoid being dominated by noise. To be consistent with the RL agent implementation, the PID control action is normalized ($a^{(pid)}_k\in[0,1]$) and multiplied by the maximum infusion rate $u$ in \eqref{eq:pk}. To avoid reset windup during induction and target changes, we implemented clamping on the integral term. The PID parameters were tuned using the Ziegler and Nichols method \cite{ziegler1942optimum} on simulations using the generic patient model (see Table \ref{tab:params}), resulting in $K_P = 9$, $K_I = 0.9$, and $K_D = 22.5$.

\section{Results}

\begin{figure}[t]
    \centering
    \includegraphics[width=0.9\linewidth]{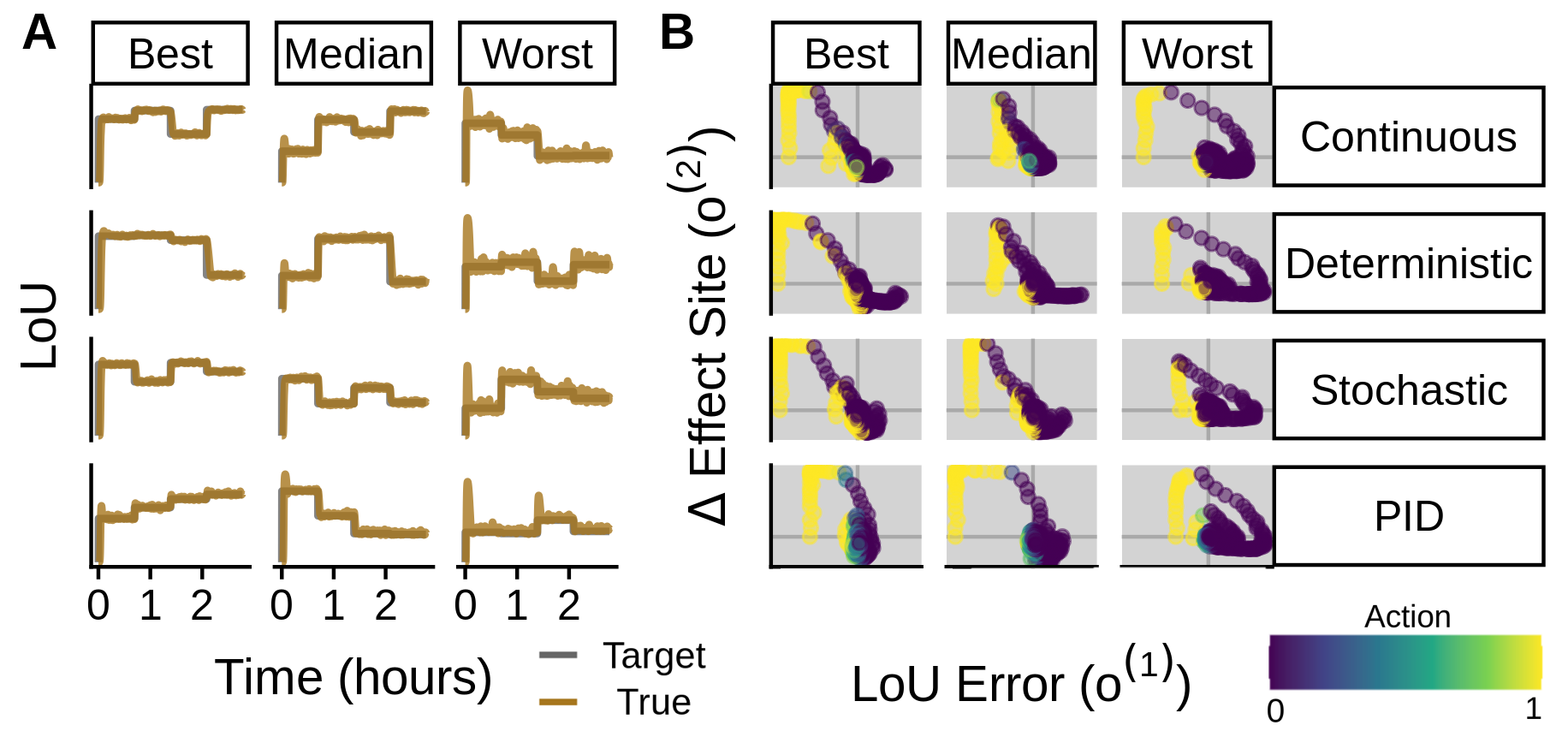}
    \caption{\textbf{A}: True and target LoU for typical/extreme cases for each controller. \textbf{B}: State-subspace trajectory for typical/extreme cases for each controller.  Each point indicates a single step in a case's trajectory.  The normalized propofol dosage administered at that step is indicated by color.}
    \label{fig:time_series}
\end{figure}

Performance was evaluated on cases that had different patient demographics and LoU targets. Figure \ref{fig:time_series}A shows sample trajectories of the true and target LoU for the cases with the worst, median, and best performance. The worst case for each controller exhibits similar increases of oscillatory behavior, especially pronounced at set-point escalations. Continuous RL used less propofol than PID during induction (186 mg $\pm$ 57 and 210 mg $\pm$ 55) and throughout a whole case (2430 mg $\pm$ 763 and 2457 mg $\pm$ 760), but more during maintenance (15 mg/min $\pm$ 5.6 and 14 mg/min $\pm$ 5.3). In the state-space view of these trajectories (Figure \ref{fig:time_series}B), it is apparent that all trajectories involve a nearly linear decision threshold with an intercept near the origin $(o^{(1)},o^{(2)})=(0,0)$. 

\begin{figure}[t]
    \centering
    \includegraphics[width=0.85\linewidth]{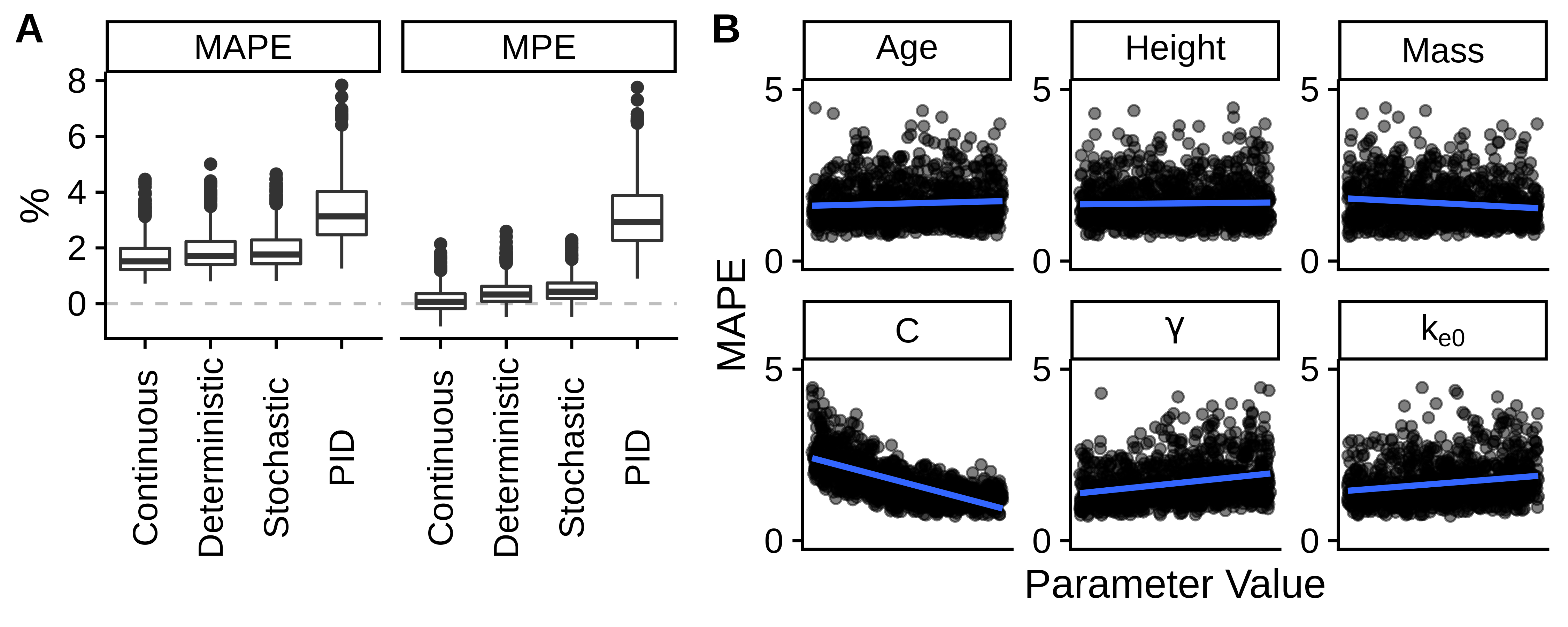}
    \caption{\textbf{A}: Median absolute performance error (MAPE) and median performance error (MPE) across 1,000 test parameterizations for each of the four controllers. \textbf{B}: Association between MAPE and PK/PD parameters for continuous action mode.  Each point represents a test-episode, positioned by that episode's performance and a PK/PD parameter.  Overlaid blue lines represent linear trend.}
    \label{fig:performance}
\end{figure}

The controller performances were evaluated using the per-episode median absolute performance error (MAPE) and median performance error (MPE), where performance error is defined as $PE_k=100\frac{y_k-y^*_k}{y^*_k}$. All RL test modes outperformed the PID controller (Figure \ref{fig:performance}A). Among the RL test modes, the continuous action mode had the best performance. Notably, the continuous mode had a median (across episodes) MPE near zero, suggesting that the ability to select continuous infusion rates helped reduce the controller's bias. On the contrary, the PID controller had nearly equivalent MAPE and MPE, suggesting that its MAPE was limited by maintaining LoU at values slightly above the target. Adjusted 2-sided paired t-tests showed that all policies have significantly different mean MPE and MAPE (p$<$0.05). The continuous RL controller was robust to variation in patient age and height, but sensitive to differences in mass and PD parameters, in particular to $C$ (Figure \ref{fig:performance}B). While this initially seems to suggest that our model performs better on patients with a higher drug requirements, it is important to note that \emph{both} $\gamma$ and $C$ affect the shape of the Hill function. As such, for the range of $\gamma$ indicated in Table \ref{tab:params}, low $C$ values correspond to steeper PD responses than high $C$ values. Sampling $\gamma$ and $C$ from a \emph{joint} distribution may reduce the apparent effect of $C$ on performance. Finally, the continuous RL controller had a duration out-of-bounds error (percentage of time at 5\% or more off target) of 6.0\% as compared with 12.4\% for the PID controller.

Figure \ref{fig:policy} shows two-dimensional cross sections of the learned policy. We see that the agent learns to transition sharply between the non-infusing and infusing actions. While the decision boundary is essentially linear in the measured error and predicted effect site concentration change, this boundary shifts to promote more infusion when the LoU has been increasing to approach the target.  

\section{Discussion}

Our experiments show that the proposed RL controllers significantly outperform a PID controller. We attribute RL's superior performance to the fact that its observation provides a much richer representation of the latent state of the system under control than PID (which only observes the error). It is worth emphasizing that we used a heuristic tuning method to optimize PID parameters on a generic patient model, and it is possible that alternative tuning methods could improve PID performance in our experiment. Nevertheless, such an extension would also involve heuristics, and the ability to incorporate robustness considerations directly into the RL training paradigm yields a considerable benefit.

\begin{figure}[t]
    \centering
    \includegraphics[width=0.82\textwidth]{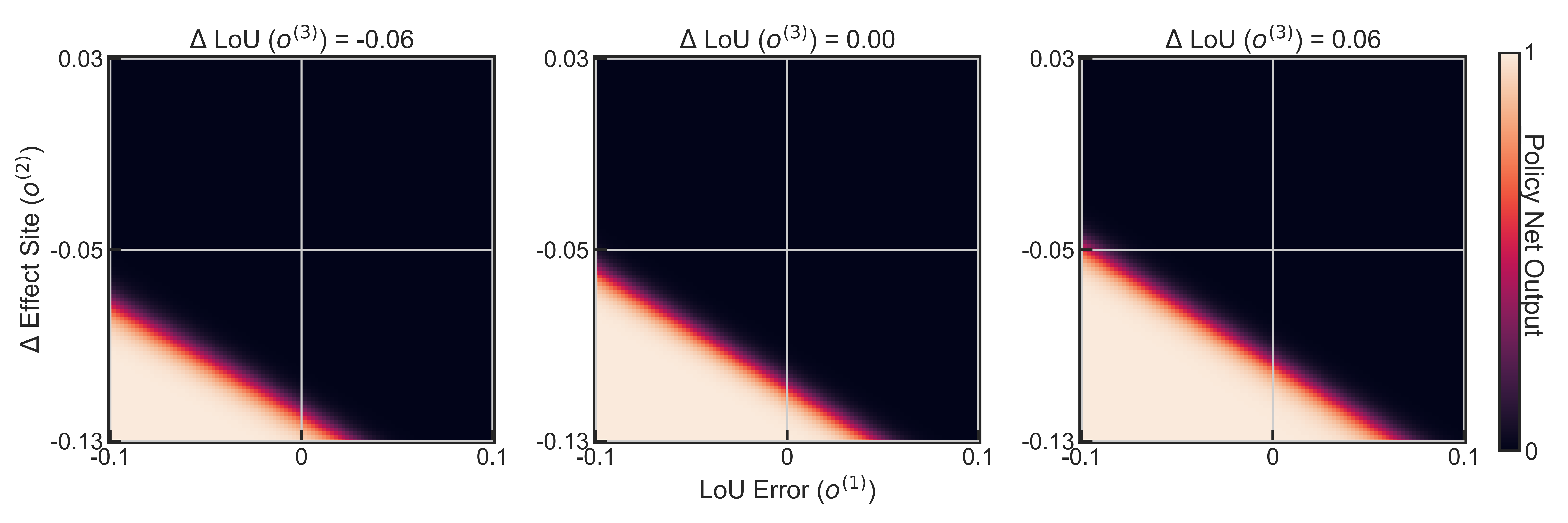}
    \caption{Policy maps show the policy net outputs ($\pi(1\mid\mbf{o}_k)$) for three different 30 second LoU changes ($o^{(3)}$) and a fixed target LoU ($o^{(4)}$) of 0.5.}
    \label{fig:policy}
\end{figure}

The behavior of the model can be interpreted by inspecting the policy. In Figure \ref{fig:policy} we see expected increases in propofol administration when the patient is further below a target concentration and when the projected effect site concentration is more rapidly falling.  The tendency to administer more propofol when the LoU has been rising may be related to behavior learned during set-point transitions or an encoding of the patient sensitivity to a given dosage history of propofol.  Given that the agent's internal generic patient model encodes the previously administered drugs, these policy maps suggest that the agent has learned the interaction between change in LoU and predicted effect site concentration change to determine at which error level drug should be administered.

The ideal way to test this algorithm would be conducting a closely monitored prospective clinical trial.  Reinforcement learning algorithms are notoriously difficult to evaluate.  Usually there is not an opportunity to collect prospective data according to the agent's policy, and policies are instead evaluated on retrospective data collected according to a different policy \cite{gottesman2018evaluating}.  In this study we changed the propofol dosage exactly every 5 seconds, whereas in standard practice dosages are changed sporadically with individual infusion rates lasting minutes to hours.  Reasonable approaches to evaluating this algorithm prospectively include non-human studies with standard dose-safety limits or clinically with a human-in-the-loop study where the agent acts as a recommender system.

\bibliography{references}
\bibliographystyle{splncs04}

\newpage
\section*{Appendix}
\begin{appendix}

The algorithm described in Section \ref{sec:training} is provided in detail in Algorithm \ref{alg:cross-entropy}. Training will terminate once either a maximum number of batches $i_{max}$ is executed or the desired batch reward $\bar{r}_{min}$ is obtained.

\begin{algorithm}
\SetAlgoLined
\KwIn{$p$, $N$,$i_{max}$,$\bar{r}_{min}$}
\KwOut{$\pi:\mc{O}\ra\mc{A}$}
 randomly initialize policy net weights $ w_\pi $ \;
 set $i=0$, $\bar{r}=-\infty$\;
 \While{$i<i_{max}$ \emph{and} $\bar{r}<\bar{r}_{min}$}{
  sample model parameters and targets\;
  simulate $N$ episodes and compute rewards $\{r_n\}_{n=1,\dots,N}$\;
  select episodes with top $p$ percentile of rewards\;
  compute cross-entropy loss between actions performed in the top episodes and associated policy net outputs $\netpolicy$\;
  perform stochastic gradient descent step to reduce the cross-entropy loss with respect to the policy net parameters $ w_\pi $\;
  set $i=i+1$, $\bar{r}=\frac{1}{N}\sum_n r_n$\;
 }
 \caption{Cross-entropy Training}\label{alg:cross-entropy}
\end{algorithm}

\begin{figure}[h]
    \centering
    \includegraphics[width=\textwidth]{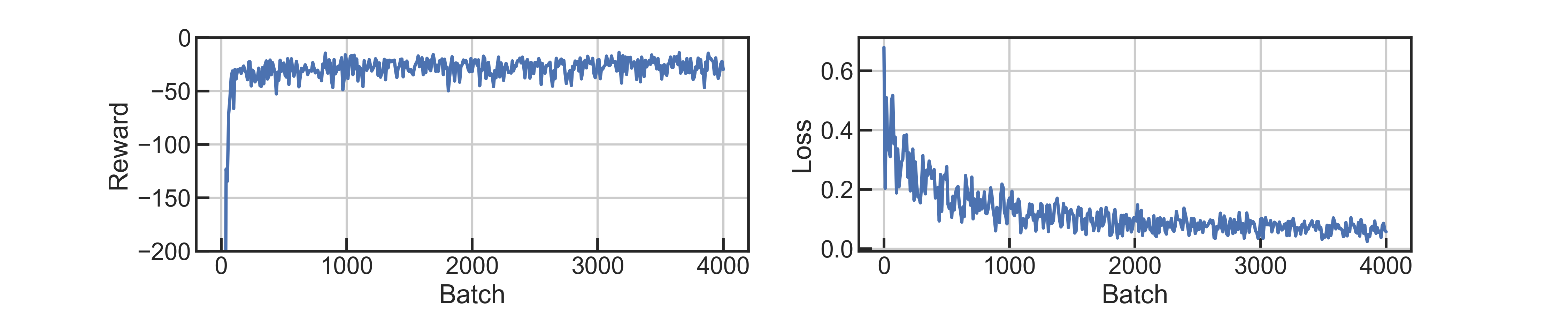}
    \caption{Round mean reward and policy network loss for each batch (corresponding to the iteration index $i$ in Algorithm \ref{alg:cross-entropy}).}\label{fig:convergence}
\end{figure}

 The per-batch mean reward $\bar{r}$ and policy network loss $\mathcal{L}$ associated with the training of our model are shown in Figure \ref{fig:convergence}, where the policy network loss is found by summing the loss over the best performing cases in the batch: $\mathcal{L}=\sum_{n\in\mathcal{N}_p}\mathcal{L}_n$ . We set the maximum number of iterations to 4,000 and visually confirmed that both the reward and loss converged. Given that the reward is represented as the negative of an absolute value, the maximum possible reward is zero, which is obtained only when the true LoU exactly matches the target LoU for the entirety of a case. Due to the inherent limitations of the environment model (for example the delay between infusion and change in effect site concentration), we expect some non-negligible error to occur at induction and target change points.

\end{appendix}

\end{document}